**Finding Variants for Construction-Based Dialectometry:**

**A Corpus-Based Approach to Regional CxGs**

Jonathan Dunn

Illinois Institute of Technology

jdunn8@iit.edu

[Short Title: Finding Variants for Dialectometry]

**Finding Variants for Construction-Based Dialectometry:**

**A Corpus-Based Approach to Regional CxGs**

*Abstract*

This paper develops a construction-based dialectometry capable of identifying previously unknown constructions and measuring the degree to which a given construction is subject to regional variation. The central idea is to learn a grammar of constructions (a CxG) using construction grammar induction and then to use these constructions as features for dialectometry. This offers a method for measuring the aggregate similarity between regional CxGs without limiting in advance the set of constructions subject to variation. The learned CxG is evaluated on how well it describes held-out test corpora while dialectometry is evaluated on how well it can model regional varieties of English. The method is tested using two distinct datasets: First, the International Corpus of English representing eight outer circle varieties; Second, a web-crawled corpus representing five inner circle varieties. Results show that the method (1) produces a grammar with stable quality across sub-sets of a single corpus that is (2) capable of distinguishing between regional varieties of English with a high degree of accuracy, thus (3) supporting dialectometric methods for measuring the similarity between varieties of English and (4) measuring the degree to which each construction is subject to regional variation. This is important for cognitive sociolinguistics because it operationalizes the idea that competition between constructions is organized at the functional level so that dialectometry needs to represent as much of the available functional space as possible.

**1. Discovering Regionally Conditioned Constructions**

Construction Grammar (CxG) views language as usage-based, with structure emerging from observed usage. If this is the case, regional varieties of a language that observe different examples of usage should differ in both their grammar (the constructions they use) and their usage of their grammar (the relative frequency of each construction). Dialectometry, on the other hand, views language as a set of features subject to spatial variation (traditionally, pronunciation features but here grammatical features) and has developed methods for quantifying this variation both in the aggregate and for individual features. Dialectometry thus offers the ability to study variation on the scale required to evaluate the claim that grammar is usage-based. This paper develops a corpus-based and construction-based dialectometry in order to examine regional variation in CxGs in English. Cognitive sociolinguistics is important here because it claims that competition between constructions is organized at the functional level. This means that construction-based dialectometry needs to represent as much of the grammar as possible in order to capture competition at the functional level.

Dialectometry measures linguistic variation in the aggregate (c.f., Szmrecsanyi, 2013; Wieling & Nerbonne, 2015; Grieve, 2016) because there are too many variants to study each independently (Nerbonne & Kretzschmar, 2013; Szmrecsanyi, 2014). At the same time, dialectometry is more than an aggregation problem because each feature, in this case each individual construction, varies in the degree to which it is spatially conditioned (i.e., externally conditioned on a spatial dimension; c.f., Wieling & Nerbonne, 2011). This relationship between individual variants and aggregate variation is becoming increasingly important as the number of known externally conditioned variants within a given domain increases (c.f., the descriptive work of Labov, et al., 2005; also Argamon, et al., 2003; Biber, 2014), making it clear that individual variants cannot be studied in isolation (Séguy, 1973; Goebl, 1982, 1984). This paper formulates a dialectometric method for measuring (i) the degree to which any individual construction in a grammar is spatially conditioned; (ii) the degree to which any individual construction is predictive of a specific regional variety; and (iii) the aggregate morphosyntactic similarity between any two regional varieties across an entire CxG.

To this end, the paper presents a method for identifying morphosyntactic variants by using construction grammar induction (C2xG: Dunn, 2017, 2018)[1] to define the features in which externally conditioned variations can occur. The paper studies regional varieties of English using two datasets: First, the International Corpus of English (c.f., Nelson, et al., 2002) is used to represent eight outer circle varieties (including Irish English as a reference point); Second, a web-crawled corpus from the Leipzig collection (Goldhahn, et al., 2012) is used to represent five inner circle varieties. A single CxG is learned using the ukWac corpus of English (Baroni, et al., 2009); this grammar is used as a feature space for finding regional variations in the usage of constructions. Within each dataset, differences between varieties are examined through three sets of experiments: First, a classifier is used to learn a model of regional variations that is validated by its ability to predict the region membership of held-out samples. Second, the model's errors and feature weights are used to quantify the similarity between regional varieties. A third experiment evaluates the regions assumed in the first two experiments in order to validate the initial set of regional varieties.

### 1.1. Previous Approaches to Finding Variants for Dialectometry

From this perspective, previous work in dialectometry can be categorized in four ways: First, most work starts with data from dialect surveys, either using raw survey data or using summarized forms of survey data (Kretzschmar, 1992; Lee & Kretzschmar, 1993; Kretzschmar, 1996; Nerbonne, 2006; Nerbonne & Kleiweg, 2007; Nerbonne, 2009; Rumpf, et al., 2009; Nerbonne & Heeringa, 2010; Wieling, et al., 2011; Grieve, 2013; Siblr, et al., 2012; Proll, 2013; Pickl, et al., 2014; Kretzschmar, et al., 2014; Onishi, 2016; Wieling & Montemagni, 2016). While these approaches represent a variety of statistical methods for performing aggregation of variants, the point here is that each starts with a dataset of collected and previously analysed (e.g., transcribed) linguistic features. They are capable of discovering variants distinctive to a given dialect, but only in a feature space that is previously defined using a fixed data collection methodology. This means that these methods cannot be applied to new variants and new corpora without beginning the process over again; this makes it difficult to incorporate new variants which may be discovered during data collection or analysis.

---

[1] The code and grammar are available from https://github.com/jonathandunn/c2xg or can be installed using pip: *pip install c2xg;* the data and a snapshot of the version of the code can be found at https://s3.amazonaws.com/jonathandunn/Finding+Variants+for+Dialectometry.zip

A second class of work, much smaller, uses a pre-defined feature space but does not rely on previously collected and annotated data (Szmrecsanyi, 2009; Grieve, et al., 2011; Pickl, 2016; Wolk & Szmrecsanyi, 2016). This corpus-based work allows for the automated or semi-automated identification of a pre-defined feature space from corpora not collected specifically for the study of linguistic variations. Thus, these methods are robust to different sources of data but do not allow the discovery of variants that are unknown in advance.

A third class of work uses an undefined feature space and does not rely on previously collected and annotated data, but does so by using only lexical bag-of-words features (Peirsman, et al., 2010; Roller, et al., 2012; Ruette, et al., 2014). By using only lexical features, fewer grammatical or structural generalizations can be found while noise from location-specific and content-specific lexical variation can obscure structural variations. First, lexical features often reflect content and not structural variants. On the one hand, while many lexical items are in variation-by-reference (i.e., using "mango" to refer to a *green pepper* or "bubbler" to refer to a *water fountain*; c.f., Geeraerts, 2010), reference in this sense is unrecoverable from a passively observed corpus. On the other hand, many externally conditioned lexical choices (e.g., "Ludlow" in Roller, et al., 2012) do not represent a preference for one linguistic variant over another but rather a difference in relevant named entities: one refers to the town "Ludlow" not because of a preference for that word but because one needs to refer to a specific place. While the set of places referred to is clearly subject to spatial variation, this is not linguistic variation.

A fourth class of work on dialectometry uses some statistical methods for learning variants, but with limitations. For example, Heeringa (2004) evaluates different methods for representing variants while still ultimately drawing data from existing linguistic atlases. Other work, for example Wieling et al. (2007) and Wieling & Nerbonne (2011), uses pre-defined features but clusters these features into spatial regions without first aggregating all features into a single measure, thus allowing a quantification of the influence of each individual feature for the overall clustering. This is an important precursor to this work because it supports both the aggregated separation of regions (here, measures of region similarity; c.f., Sanders, 2007, 2010 for measures of aggregate syntactic similarity) while also maintaining individual features and measuring the degree to which they are spatially conditioned.

### 1.2. Previous Approaches to Regions and Boundaries in Dialectometry

This work differs from previous approaches to dialectometry in that it assumes the geographic space of dialects (evaluated in Section 4) but does not assume the feature space of variants. Some previous work (i.e., Nerbonne, et al., 2008; Grieve, 2014) uses clustering methods to identify dialect regions and the borders between them given a fixed set of features. These methods focus on the discovery of specific regions while this paper assumes fixed regions and focuses on the discovery of variants within them. A classification approach is taken to this problem: given a fixed set of regions and a high-dimensional learned CxG representing each region, which features contribute to distinguishing between these regions? Clustering methods face a validation problem because the output clusters cannot be evaluated against a known ground-truth.

That is not a concern here because the classification approach comes with a ground-truth for validation: the dialect model is only as good as its ability to predict the region membership of new held-out observations. In other words, the classifier makes a prediction about which features are in spatial variation and these predictions are tested by applying them to new observations. This validation step is essential given the goal of finding new variants: a model capable of predicting region membership given only information about construction usage provides evidence that this usage is, in fact, spatially conditioned.

Another strand of dialectometry (Kretszchmar, 1992, 1996; Goebl, 2006) focuses more specifically on finding dialect regions and the boundaries between them given survey-collections of known variants. These approaches force us to ask an important question: if a different set of regions had been assumed for each dataset, how consistent would the dialect model have been? This is important because methods which assume a fixed set of features must be asked to evaluate the impact that assumption has on the final model; in the same way, this paper must evaluate the impact that the assumed dialect areas have on the final choice of features. While the specific borders of the dialect regions are not altered (in part because the regions are not contiguous), this question is explored further in Section 4.

### 1.3. Dialectometry and Cognitive Sociolinguistics

The basic argument of this paper is (i) that the CxG paradigm expects regional variation in both grammar and usage while (ii) there are too many spatially conditioned variants in language to study each individually. Thus, dialectometry has taken up the task of accounting for the full space of potential variants within a given scope (here, CxGs). This is because the linguistic distance between regional varieties should capture the total variation across all available constructions. This paper uses corpus-based dialectometry to study regional CxGs in English. This whole-grammar approach to variation represents an important advance for both dialectometry as the study of variation and also for CxG as a usage-based paradigm.

Because CxG is usage-based some constructions are hypothesized to be more entrenched in a speaker's mind than others. One explanation for entrenchment is frequency: more frequently encountering a particular construction will lead a speaker to grammaticalize that form-meaning combination (c.f., Bybee, 2006). On the one hand, a construction can be more entrenched than other competing constructions, perhaps resulting from more frequent use (c.f., Langacker, 1987). On the other hand, in dialectometry frequency of usage is an indicator of a preference for a particular item: when other factors are controlled for, constructions that are used more frequently in one regional variety have been selected by that variety.

Here we have two concepts: entrenchment (or grammaticalization) as a cognitive property of speakers and frequency of usage as an observable property of constructions in corpora. How are they related? Psycholinguistic and corpus-based evidence sometimes fit together nicely (c.f., Divjak, et al., 2016) and sometimes are difficult to synthesize (c.f., Dąbrowska, 2014). Part of the problem is that most corpora (and certainly the large corpora used in this paper) are sampled from tens of thousands of individuals while psycholinguistic studies are individual-specific. For example, it seems to be the case that language learners do not converge on precisely the same grammar (c.f., Dąbrowska, 2012); this claim is not controversial from the perspective of dialectometry because we see widespread variations in grammar and usage across geographic regions. What it means is that corpus-based measures and models are operating on heterogenous data: in addition to regional variations, the CxG in this study is subject to individual variation and social variation and register variation.

How can we synthesize corpus-based variation with a plausible psycholinguistic model of individual differences within a cognitive sociolinguistics framework that is, above all, focused on meaning? First, in methodological terms we need to use modeling techniques that allow for counter-factuals (c.f., Zenner, et al., 2012) and that are validated against held-out data (c.f., Divjak, et al., 2016). Raw measures such as frequency are not sufficient in and of themselves to represent trends in a corpus (c.f., Szmrecsanyi, 2016). In the case of the entrenchment of constructions, many factors can increase the frequency of construction usage. The output of a validated model is required to be sure that these differences are meaningful. Validating corpus-based models on held-out testing data, for example through cross-validation, is equally important because it ensures the robustness of the model. These methodological points are necessary to validate corpus-based results because we need to be sure that the variations in frequency we observe are, in fact, predictable across different sub-sets of the corpus.

This still leaves the question of how corpus-based dialectometry relates to a meaning-based cognitive sociolinguistics with psycholinguistic plausibility. One approach to psycholinguistic plausibility is to use a learning algorithm such as Naïve Discriminative Learning (Baayen, et al., 2011) that is designed to mimic human learning processes and thus can be used to simulate human learning over corpus data (Milin, et al., 2016). This would replace a purely practical classifier (i.e, the Linear Support Vector Machine used here) which does not purport to model human learning but rather excels at finding the best solution, in this case a weight for each construction representing its spatial conditioning. This approach is not relevant to dialectometry, however, because it is not claimed that naïve humans possess the ability to distinguish between these varieties given only CxG usage; for one thing, no individual is likely to have observed a sufficient amount of English use across all the varieties studied in this paper to support such judgements. Distinguishing between regional varieties is not necessarily a cognitive ability. We do expect, however, that usage-based entrenchment across different communities using English will result in precisely these sorts of regional variations. Thus, dialectometry is not modeling a cognitive ability so much as modeling the effects of a cognitive and social process.

The commitment to language as cognition "encompasses shared and socially distributed knowledge and not just individual ideas and experiences" (Geeraerts, 2016: 533). In other words, a phenomenon like regional variations in usage does not need to be contained only within individual speakers in order to be relevant to cognitive linguistics. The idea that grammar is usage-based, with some constructions more entrenched (and thus more productive) as a result of more frequent observation, predicts that we will observe regional CxGs (c.f., the Entrenchment-and-Conventionalization Model;  Schmid, 2016). Constructions that are more entrenched are used more frequently which, in turn, makes them more entrenched in a given speech community (for example, Hollmann & Siewierska, 2011). This link between entrenchment and social variation needs to be investigated and a corpus-based approach is the only feasible method given the number of individuals involved.

The problem for corpus-based approaches to CxG variation is that the competition between constructions is at the level of function or meaning. For example, when studying variations in lexicalization (Zenner, et al., 2012) we must organize variants in an onomasiological fashion, so that lexical items that represent the same concept are quantified by their relative share in lexicalizing that concept. In the same way, constructions express specific meanings, carrying out specific functions, so that competition between constructions is organized around the total functional load of a grammar. In the past, studies of variations in construction usage have focused on a very small number of constructions that are selected specifically because they are known to have overlapping functions (i.e., Levshina, 2016). This extreme limitation in features is not acceptable from a dialectometric perspective because it makes an arbitrary selection of features based on convenience.

From a cognitive sociolinguistic perspective, there may be many unknown or unstudied constructions that overlap with the functional load of those constructions chosen for a particular study. These constructions are relevant but not included in the analysis. The question from both dialectometry and cognitive sociolinguistics, then, is simple: how do we capture as much of the choice space of possible variants (dialectometry) or possible functions (cognitive sociolinguistics) that are available to speakers? CxG induction provides the ideal selection method because it captures that largest number of variants and represents as much of the language's functional load as possible.

On the one hand, no grammar is perfect and the claim is not being made that the grammar discussed in Section 2 captures all possible structures in all varieties of English. On the other hand, this grammar does represent significantly more of the grammar than any previous study of CxG variation or dialectometry. In terms of cognitive sociolinguistics, this means that the models of regional variation in Section 3 are able to represent variation across a significant number of constructions with overlapping functions. This is essential for studying CxG dialectometry: we must avoid arbitrarily selecting a limited set of features and instead capture as many ways of expressing meaning as possible.

**2. Construction Grammar Induction to Discover Potential Variants**

CxG views language as a set of symbolic form-meaning mappings (Goldberg, 2006; Langacker, 2008). There is a large body of research using CxG to represent linguistic structure (e.g., Kay & Fillmore, 1999) and studying variation and change using CxG representations (e.g., Claes, 2014; Hoffman & Trousdale, 2011; Hollmann & Siewaierska, 2011; Goldberg, 2011; Gisborne, 2011; Uiboaed, et al., 2013). The advantage of CxG from the perspective of dialectometry is that, while constructions do contain relations between their internal slots, they remain countable entities as a whole. Thus, an individual construction represents a single discrete structure in a way that the atomic units of other grammatical formalisms do not. This is important because it provides a straightforward way to vectorize the usage of a CxG from passive observations: each individual construction is a feature (column) and each observation of a construction increases that feature's frequency of usage. Because each construction represents a specific choice, insofar as one construction has been used instead of another, this means that morphosyntactic preferences are easily quantified within the CxG paradigm (this idea of usage revealing morphosyntactic preferences is explored further in Section 3.3).

For example, a construction could be a schematic form (e.g., the ditransitive in 1a, below) or an item-specific form (e.g., the ditransitive with the verb "give" and the object "hand" in 1b). This is important because it helps to distinguish which choices are in competition: (1b) competes with the alternate form in (1c) in a way that the more schematic and non-idiomatic ditransitive in (1a) does not. The advantage of CxG is that it can represent usage in such a way that structures can be quantified at the level at which they are in competition with other structures.

(1a) John sent Mary a letter.

(1b) John gave Mary a hand.

(1c) John helped Mary.

Further, CxG combines lexical, syntactic, and semantic representations to create robust grammatical generalizations; this is important for representing morphosyntactic choices at precisely that level which is in variation. For example, (2a) and (2b) below share the same dependency structure at a purely syntactic level but it is clear that these forms are not in competition: (2a) is in competition with (2c) while (2b) is not. This is a case in which lexical and semantic representations in addition to syntactic representations are necessary for quantifying morphosyntactic preferences. The idea of competition between constructions is similar to the problem of preemption: why are some constructions and not others used in specific linguistic contexts (c.f., Stefanowitsch, 2011, Goldberg, 2011)? Does the entrenchment of one construction prevent the use of competing constructions? In terms of dialectometry, if a construction is particularly entrenched in a regional variety it is externally conditioned on a spatial dimension, thus adding non-linguistic factors to the problem of preemption.

(2a) John gave his neighbor a piece of his mind.

(2b) John sent his neighbor home  in his car.

(2c) John told his neighbor off.

The difficulty of CxG for dialectometry is to define the feature space: which constructions should be present in the grammar and which should be examined in a given study? How will these constructions be identified in large corpora while maintaining reproducibility? The essential problem is that there are many potential constructional representations that have been observed in usage but only some are productive for speakers of a particular variety. This paper uses construction grammar induction (Dunn, 2017, 2018) to overcome both difficulties: first, to learn a CxG from an independent corpus and, second, to extract these constructions from regional corpora in order to quantify their frequency of use. The frequency of construction usage becomes the feature space for dialectometry. This section briefly presents the CxG induction algorithm in order to characterize the nature of the CxG representations used for dialectometry and to evaluate the specific CxG used in this study.

### 2.1. Representing CxGs

The basic idea of CxG is that grammar is more than just a formal system of stable but arbitrary rules for defining well-formed sequences. Rather, grammar consists of meaningful constructions in the same way that a lexicon consists of meaningful words. This brings together two important premises: First, that grammar consists of meaningful symbolic units (e.g., Langacker's Cognitive Grammar); Second, that co-occurrence and distribution are indicators of meaning (e.g., Firth, 1957; Cilibrasi & Vitanyi, 2007). These premises suggest that constructions, like words, can be studied and defined as a set of co-occurring elements in a corpus. Co-occurrence is a measure of the relative productivity of competing representations; for example, we expect the more generalized constituent representation in (3c) to co-occur more significantly than the single unit representation in (3b) because there are many possible configurations like (3b) that are covered by the representation in (3c). We expect grammatical constructions to display internal co-occurrence that distinguishes them from unproductive representations.

(3a) "Bill gave his neighbor a piece of his mind."

(3b) [NN – VB – PRN – NN – DT – NN – PREP – PRN – NN]

(3c) [NP – VB – NP – NP]

(3d) [NP <ANIMATE> – VB <TRANSFER> – NP <ANIMATE> – NP]

(3e) ['a piece of his mind']

The constructional representations learned are sequences of slots, as in (3b) through (3e), each slot constrained by syntactic, semantic, or lexical restrictions at the word- or constituent-level. For example, the slot VB-PHRASE <TRANSFER> can be filled or satisfied by any verb constituent from that particular semantic domain (e.g., "give", "send", "sell"); this means that the observed linguistic expression in (4b) satisfies the slot requirements of the construction in (4a) and counts as an instance of that construction. Thus, this construction is defined in terms of both purely syntactic information (e.g., NN-PHRASE) and semantic selectional constraints (e.g., VB-PHRASE <TRANSFER>). Note that individual syntactic units are indicated by small caps (e.g., NN).

Semantic selectional restrictions are represented using domains enclosed in brackets (e.g.,

<ANIMATE>). Lexical items are represented using single quotation marks.

(4a) [NN-PHRASE – VB-PHRASE <TRANSFER> – NN-PHRASE <ANIMATE> – NN-PHRASE]

(4b) "The child gave his brother a new book."

(4c) ['give' – NN-PHRASE <ANIMATE> – 'a break']

(4d) "Please give me a dollar."

(4e) "Please give me a break."

Constructions are posited at multiple levels of abstraction, so that more schematized representations like (4a) co-exist with item-specific representations like (4c). In this case, (4c) is a partially-fixed instance of the ditransitive that is not fully compositional. Thus, the linguistic expressions in (4d) and (4e) are produced by separate but related constructions that differ in their level of abstraction. The output representations are called constructions and all linguistic expressions described by a construction are called its constructs. Constructions are represented as sequences in which each unit or slot is constrained at the syntactic, semantic, or lexical level. All sequences that meet the constraints posited by a specific construction count as instances of that construction. Thus, for dialectometry, the representation in (4a) counts as a single feature or variant and the occurrence of (4b) in a sample counts as an observation of that variant.

In this implementation, the lexical level consists of word-forms as indicated in the orthography by whitespace. The syntactic representation of word-forms uses a part-of-speech tagger: here, RDRPosTagger (Nguyen, et al., 2016; the tag-set is from the Universal POS tag-set; Petrov, et al., 2012), the only supervised component of the algorithm. The word-level semantic representation is learned from the input corpus using word2vec (with 500 dimensions quantified using a skip-gram model; GenSim, Rehurek & Sojka, 2010) together with K-Means clustering (k = 100) to produce a dictionary providing domains for each word. Slots in a construction can be filled by constituents as well as by individual lexical items. A context-free constituent grammar is learned by finding purely syntactic constructions, assigning these constituents to a head unit, and allowing them to fill individual slots. Such slot-filling constituents are assigned to the semantic domain of their head unit.

The essential idea of grammar induction is that if we can evaluate the quality of a grammar against an unannotated corpus then it is possible to search over potential grammars until the best one is found. The problem, then, is to develop a measure of grammar quality that does not require an annotated corpus. The details of this problem are outside the scope of this paper; more information is provided in the external resources and in recent work on computational construction grammar (Dunn, 2017, 2018); an evaluation of the stability of the grammar used in this paper is contained in the external resources.

## 2.2. Representative Examples of Learned Constructions

To illustrate the type of constructions present in the grammar, examples of constructions are shown in (5) through (11). For each construction, several utterances are given that are examples or manifestations of that construction. For the dialectology experiments below, each construction is a feature (i.e., a variant: 5a) and each manifestation of a construction is an observation of that feature (i.e., contributes to its frequency in a given sample: 5b through 5e). The first example, in (5a), is a modified adverb construction in which adverbs are modified to include information about vagueness. For example, the difference between "at one o'clock yesterday" and "at about one o'clock yesterday" is the certainty of the expression.

(5A) [ADVERB – 'about']

(5b) "at about"

(5c) "how about"

(5d) "only about"

(5e) "on about"

The second example, in (6), is an argument structure construction with a verb-specific direct object. Here the verb is constrained to a specific lexical item ("provide") and the direct object is constrained to a semantic domain (which is unlabelled because of the way semantic domains are created from word embeddings). These semantic domains are somewhat opaque to introspection because they are formed in a bottom-up fashion. Nonetheless, this example shows the importance of multiple levels of abstraction for CxGs because we do not know in advance which representations will best describe the language observed in a corpus. This also shows the level of usage that is made available for dialectometry.

(6a) ['provide' – <25> – <25>]

(6b) "provide added value"

(6c) "provide an opportunity"

(6d) "provide general advice"

(6e) "provide information about"

Becoming more complex, the example in (7) represents an event phrase that contains both a main verb (e.g., "want,") as well as an infinitive verb (e.g., "improve"), both of which are defined semantically. This, again, creates a somewhat opaque representation (in 7a) but the generalizations represented by the instances of this representation (7b through 7e) are not opaque. These are all cases in which the first verb encodes the intentions of the actor and the second verb encodes the action that has been taken.

(7a) [<25> – 'to' – <14>]

(7b) "designed to ensure"

(7c) "want to improve"

(7d) "made to ensure"

(7e) "able to understand"

Moving to a more complex verb phrase, the example in (8a) describes a basic verb phrase (e.g., "to consider how") embedded within an evaluative verb describing how the speaker perceives that event. This is similar to the example in (7a) but offers an interesting contrast in two ways: First, here the main verb is represented syntactically. This means that this is a more open slot than the first position in (7a) and can be filled by a much wider range of lexical items.

Second, the syntactically-defined final slot provides an adverbial particle for constraining the embedded verb. This example shows how the specificity of slot constraints influences the range of instances that are described by a construction.

(8a) [VERB – 'to' – <25> – ADVERB]

(8b) "need to consider how"

(8c) "wish to consider how"

(8d) "want to be here"

(8e) "like to find out"

The grammar also contains complex noun phrases as in (9a). This construction encodes a noun phrase with a modifying prepositional phrase. Most of the noun phrase is defined syntactically, so that it has relatively open slots. The final slot, however, is defined by its semantic domain. This has the practical effect of limiting the main noun to those that can be modified by items from this particular domain.

(9a) [DETERMINER – NOUN – ADPOSITION – <14>]

(9b) "some experience in research"

(9c) "a need for research"

(9d) "the process of planning"

(9e) "a number of activities"

The grammar also captures clause-level structure, as in the subordinated noun phrase in (10a). This construction provides syntactically-defined noun phrases that are attached to main clause verbs and act as the subject for additional modifying material that remains unspecified. We can think of this as a linking construction in the sense that, when it is attached to part of an argument structure construction, it allows sub-ordinate clauses to become arguments.

(10a) [SUBORDINATE-CONJUNCTION – <25> – ADJECTIVE – NOUN]

(10b) "whether small independent companies"

(10c) "that the international community"

(10d) "because the current version"

(10e) "while the other party"

Finally, the example in (11a) describes a partial main clause and represents the largest constructions currently identified by the algorithm. This is not a complete clause in that it must be joined together with an additional noun phrase construction. The point of these examples has been to show the sorts of constructions present in the grammar and the sorts of instances that, when observed, contribute to the frequency of the construction as a feature for dialectometry. It is important to note that, in the current implementation, constructions cannot fill slots in other constructions. Future work will add another pass to the algorithm in order to produce larger sentence-level and clause-level constructions that are composed of smaller constructions.

(11a) [PRONOUN – AUXILIARY-VERB – VERB – PARTICLE – <25>]

(11b) "you should continue to receive"

(11c) "i was told to make"

(11d) "they were going to have"

(11e) "this was going to be"

There are 4,504 constructions in the grammar. A descriptive breakdown by length (number of slots) and representation (type of slot constraint) is shown in Figure 1. First, on the left we see that the distribution of construction lengths is relatively even. The peak type frequency is with constructions containing 3 slots (multiple lexical items may fill a single slot within particular instances of a construction). Note that horizontal pruning (c.f., Wible & Tsao, 2010) is used to remove constructions from the grammar that are entirely contained within other constructions; this favors longer constructions when there is overlap. The grammar contains a total of 15,300 slot constraints across all constructions; these are broken down by type of representation on the right. The largest category of representations, at 43.1%, is syntactic. This is expected because many grammatical patterns can be described in purely syntactic terms. The smallest category of slot constraints is lexical, at 26.9%. This is also expected because item-specific representations, while important to capture, do not provide as much generalization as syntactic and semantic representations. A further examination of the constructions used for dialectometry and raw regional variations in those constructions is available in the external resources accompanying this paper.

*Figure 1. Descriptive Breakdown of Constructions in Grammar*

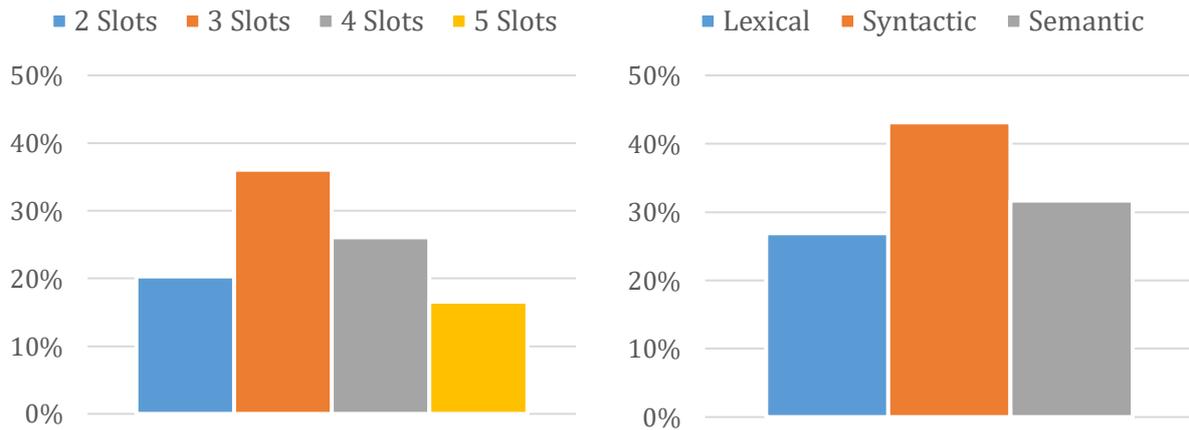

## 3. Dialectometry Through Classification

We take a classification approach to dialectometry: given a set of observations from known regions, what set of features can best distinguish between these observations? In this experiment, CxG features learned in Section 2 are applied to the task of distinguishing regional varieties. The classification problem provides three pieces of information: (1) the accuracy of predictions, representing how well the feature set is able to predict the regional variety which a particular sample comes from; (2) the relative importance of all features in making that prediction and those features most predictive of (and thus unique to) a given region; and (3) the similarity between regional varieties given both feature weights and classification errors. This section assumes fixed regions (i.e., that Indian English comes from a given political nation-state), but this set of regions is itself evaluated in Section 4.

### 3.1. Datasets

The first source of regional varieties of English is the International Corpus of English, with observations from eight outer circle varieties: East Africa, Hong Kong, India, Ireland,

Jamaica, Nigeria, the Philippines, and Singapore. The use of these varieties reflects the fact that these regional Englishes are an important part of dialectology: there is a continuum on which dialects, regional varieties (c.f., Kortmann, et al., 2004; Schneider, 2007), heritage varieties (c.f., Nagy, 2016), and non-native varieties (c.f., Henderson, et al., 2013) are different ways of framing the same phenomenon in either linguistic, spatial, political, or educational terms. To generalize, this paper prefers the term *regional variety*. For each of these areas, the respective ICE corpus is divided into chunks of 2,000 words without regard to register (i.e., spoken or written language) or topic (i.e., newspapers or business letters). The idea is to randomly distribute observations of a given region across registers and topics so that any generalizations that remain are specific to the region in question. Thus, any individual sample may contain text from multiple registers and multiple speakers. Taken together with the balanced collection methods for the ICE corpus, this ensures that the model does not rely on register-specific features. Each region contains between 500 to 700 samples, some of which (795 samples across all regions) are used as a development corpus.

The second source of regional varieties of English is web-crawled corpora from the Leipzig corpora collection (Goldhahn, et al., 2012) with observations from five inner circle varieties: Australia, Canada, New Zealand, United Kingdom, and South Africa. Note that different regions in this dataset were collected in two different years: 2002 (AU, CA, NZ, UK) and 2011 (ZA). These corpora are presented as individual sentences and samples are formed by randomly aggregating sentences into samples of 2,000 words. Unlike ICE, these corpora are not manually collected and, as a result, are not balanced according to topic or register. On the other hand, the corpora are collected using the same methods and represent the same diverse types of web-crawled language use. This corpus is much larger, containing between 8,700 and 9,400 samples per region, with the development portion containing 6,710 samples. Variations in register are controlled for by (i) the large number of samples and (ii) the aggregation of randomly selected sentences from a region into samples. Because of differences in register, these two data sets are not directly compared in this paper; the goal is to model regional varieties while holding register constant in order to ensure that regional varieties are the only source of variation.

### 3.2. Models

A Linear Support Vector Machine classifier is used to learn feature weights (Joachims, 1998; c.f., Dunn, et al., 2016, in reference to using Linear SVMs for finding variants). This is a supervised method that observes a number of samples (i.e., vectors of construction frequencies representing samples from a given region) and estimates a function for mapping that vector into a hyperplane maximizing the separation between classes (i.e., regions). A Linear SVM is preferable to other linear classifiers with inspectable feature weights, such as Naïve Bayes, because it can better handle redundant representations. This is important because constructions vary in their level of abstraction so that a single utterance may have several constructions describing it, producing correlated features. The search space is high-dimensional (with 4,504 dimensions, given the grammar); this is because we start with the assumption that any element in the grammar can be spatially conditioned. This is a much higher-dimensional problem than existing clustering-based methods in dialectometry but much lower-dimensional than traditional text classification problems. Dimension reduction is not used because it would only serve to disguise the importance of individual features.

Constructions are quantified using their raw frequency; since all samples are the same size, this is relative frequency. Thus, the grammar is turned into a vector that contains the frequency of each construction in each observed sample. It is important to note that these feature vectors differ from the feature weights discussed shortly, which represent the preference of a specific region for a specific construction. Feature weights are properties of the learned classifier (the Linear SVM) and their validity follows from the validity of the classifier's predictions. Thus, the feature frequencies are observed (as the input to the classifier) and the feature weights are predicted (as the output of the classifier).

Because this is a supervised method, the validity of the results depends heavily on how we segment the data: the observations used to choose the model parameters (development data) must be kept separate from the observations used to estimate a mapping function (training data) and the observations used to evaluate the model's predictions (testing data). Following best practices, we further employ cross-validation to randomly iterate over different training and testing segmentations in order to ensure that the evaluation does not depend on a limited segment of testing data. Within this experimental paradigm the specific classifier used is less important than the validation of the estimated model on held-out test data. In other words, the classifier (here, a Linear SVM) generates and chooses a specific hypothesis about the spatial conditioning of each construction given the training data. There are many reasons why this hypothesis can fail: the classifier itself may perform poorly in estimating a model, the features may simply not be in significant spatial variation, or the grammar from Section 2 may fail to represent regional usage. But, given sufficient controls, there is only one reason why the classifier can make accurate predictions: because the vectors it is given represent structures that are subject to predictable spatial variation.

A grid-search for optimum parameters and normalization methods is performed using a randomly selected development corpus; reported classifier performance is computed using 10-fold cross-validation. Each region in these datasets is approximately the same size; limited over-sampling (minority classes) and under-sampling (majority classes) ensures balanced classes.

### 3.3. Selection Signatures

Morphosyntactic dialectometry in this paradigm depends on the fact that speakers have a large number of grammatical structures available to them but can only choose a small sub-set of these structures in actual usage. Positive evidence for a speaker's preference is provided by each observed structure and negative evidence by each unobserved structure. In terms of cognitive sociolinguistics, an entire CxG can perform all of the functions the language is used for. Studying only a few constructions in isolation limits the functions that are represented. Thus, even if constructions are chosen because they have overlapping functions (i.e., Levshina, 2016), this approach (i) may miss constructions that fulfil those same functions in other contexts or (ii) may miss some functions that are covered by those constructions in other contexts.

So long as the total choice space is relatively well covered (i.e., so long as the CxG has descriptive adequacy), the amount of negative evidence will be much higher than the amount of positive evidence. Corpus-based dialectometry does not require the active elicitation of either specific variants or specific minimal pairs: given enough passively observed language use, the observed frequency of each structure (the input to the model) supports the estimation of each region's preferences for that structure against its competition (the output of the model).

We call these output representations for each region their *selection signatures*. The basic idea is (i) that speakers have a very large number of choices available to them, (ii) that speakers can only choose a limited number of structures in actual usage, (iii) that varying preferences for a given structure across regions is a matter of externally conditioned variation, and (iv) that observations of the entire choice space support the quantification of regional preferences. In terms of cognitive sociolinguistics, so long as the CxG covers a significant number of the functions or meanings that the language can be used to express, these selection signatures quantify regional variation in how functions are expressed across the entire grammar. This assumes that the dataset is sufficiently large and homogenous that samples from different regions represent the same general inventory of functions. For example, this would pose a problem if texts from Singapore English were entirely of a religious nature and none of the texts from other varieties were of a religious nature.

Given observations sampled from Region A and Region B, this selection signature provides a quantification of the morphosyntactic preferences represented by these samples. On the one hand, the relative preference for a specific structure within this choice space has many language-internal causes and is not relevant here. On the other hand, external conditioning in this paradigm is reflected by differences in the relative preference for a given structure across different regions. In the extreme case, a regional variety of a language may not contain a specific structure at all, so that its preference for that structure is zero (c.f., Szmrecsanyi, 2016). This is a case of a regional variety having a unique grammar. More commonly, however, regional varieties display subtle differences in usage preferences. In these cases, a regional variety contains a given structure in its grammar but varies in its relative preference for that structure. In this work, a Linear SVM is used to estimate selection signatures over many samples from each region.

### 3.4. Measuring Model Validity

The first question is whether the classifier is able to learn the relative spatial conditioning of each construction in the grammar: true positives occur when the model assigns unseen samples to the correct region and false positives occur when the model incorrectly assigns a sample to a given region. The standard measures used to evaluate such an experiment are precision (the proportion of predictions for region X that actually belong to region X, or $TP / (TP + FP)$ where $TP$ is a true positive) and recall (the proportion of samples from region X that were correctly classified, or $TP / (TP + FN)$ where $FN$ is a false negative). The F-Measure reported here is the harmonic mean of these two measures averaged across all classes. Precision and Recall are reported even though they are the same as F-Measure in order to show that high performance on majority classes does not falsely inflate the overall F-Measure. This is true in the aggregate, but is not true for each region individually as shown in Figures 2 and 3. The results, in Table 1, show that this approach makes predictions that are quite accurate on held-out data: an F-Measure of 0.92 on the smaller ICE corpora and an F-Measure of 0.97 on the larger web-crawled corpora. This validates the model because a sample's provenance can be predicted given only a vector containing the observed frequency of constructions. The majority baseline is provided to show that these high accuracies do not simply result from imbalanced classes.

*Table 1. Performance of CxG-based Region Classification*

|  | Precision | Recall | F-Measure | Majority Baseline |
|---|---|---|---|---|
| Inner Circle Varieties | 0.97 | 0.97 | 0.97 | 0.07 (F1) |
| Outer Circle Varieties | 0.92 | 0.92 | 0.92 | 0.03 (F1) |

This high accuracy licenses further investigations into what supports these predictions and what these predictions can tell us about regional varieties of English. Unlike other statistical methods used for dialectometry, such as logistic regression, a classifier does not produce significance values for its predictions. Intuitively, however, it is a higher threshold to say that a feature set can accurately predict region membership on held-out data than to say that a logistic regression analysis shows a significant relation between a feature and a given regional variety. In other words, traditional approaches to linguistic variation determine the strength of the relationship between an individual feature and a given region. The present approach first determines how a large number of individual features aggregate to distinguish between regions and then defines the strength of the relationship between individual features and each region using the importance of each feature for this aggregate model. This allows us to quantify spatial conditioning as a group phenomenon operating over many individual constructions while still measuring the conditioning of individual constructions.

More detailed results for outer circle varieties are shown in Figure 2. First, we notice that Irish English performs the best, reaching almost perfect accuracy. This goes along with our intuitions that in historical terms the formation of Irish English is different from these other regional varieties (and, of course, the term outer circle is being used somewhat loosely to include Irish English). While all regions perform moderately well, we can also learn about the model through the errors it makes: Jamaican and Singapore English are the most commonly misclassified varieties, so that they have lower F-Measures. They differ in the source of their errors, however: Jamaican English has lower recall, which means that samples of Jamaican English were predicted to belong to other varieties; Singapore English has lower precision, which means that samples of other varieties were predicted to belong to Singapore English.

*Figure 2. Results By Region, International Corpus of English*

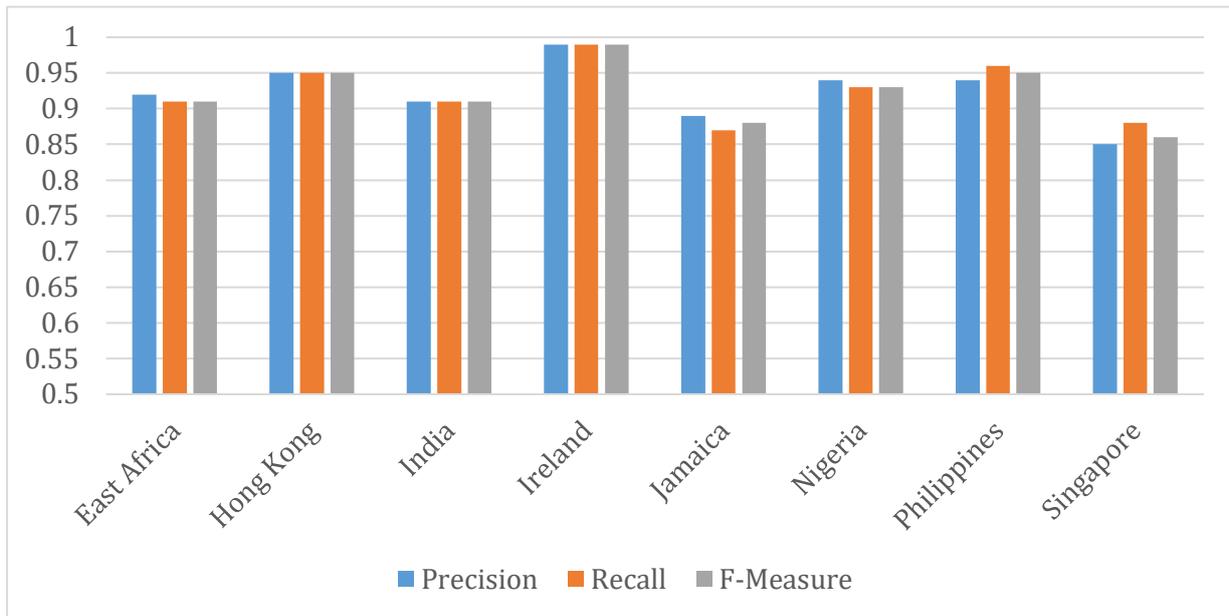

The same analysis for inner circle varieties is shown in Figure 3. The overall performance here is a bit higher and this is reflected in the fact that no variety falls below 0.95 F-Measure. The lowest performing varieties are Australian and New Zealand English, which we will investigate further below using error analysis: why does the model make more errors for these varieties?

*Figure 3. Results By Region, Leipzig Corpora*

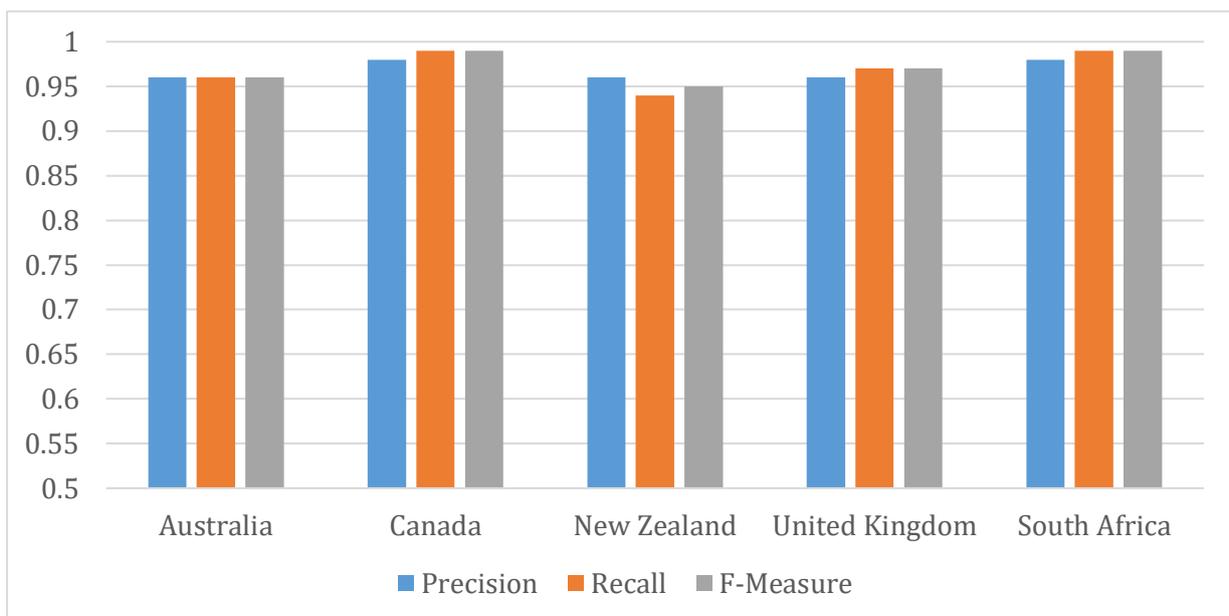

We represent errors using the confusion matrix shown in Table 2. Each row represents samples from a given region and each column represents predictions for these samples. True positives occur when the row and column have the same region; these are shown in bold. The numbers represent individual samples from each region. For example, observations of East African English were correctly predicted 513 times but confused most commonly with Indian English (18 times) and Jamaican English (12 times). The overall percentage of errors is low, but the distribution of errors can be used to reveal which varieties the model commonly mistakes (darker shading indicates a higher number of errors). For example, Irish English is lightly shaded throughout, with only 4 samples wrongly predicted to be from another region. Thus, it is relatively well-described. Jamaican English, on the other hand, has errors with a number of regions, especially East African and Singapore English. This suggests that these regions, with more misclassifications, are more similar. This is explored further below.

*Table 2. Confusion Matrix for Classification with ICE*

|       | E. AF | HK  | IN  | IR  | JA  | NI  | PH  | SI  |
|-------|-------|-----|-----|-----|-----|-----|-----|-----|
| E. AF | **513** | 1 | 18 | 0 | 12 | 10 | 1 | 10 |
| HK    | 2 | **536** | 0 | 0 | 5 | 3 | 5 | 14 |
| IN    | 10 | 3 | **518** | 0 | 10 | 3 | 5 | 18 |
| IR    | 0 | 0 | 0 | **557** | 0 | 0 | 4 | 0 |
| JA    | 15 | 7 | 11 | 0 | **483** | 10 | 6 | 24 |
| NI    | 9 | 0 | 1 | 0 | 12 | **522** | 5 | 15 |
| PH    | 0 | 6 | 3 | 4 | 5 | 0 | **554** | 4 |
| SI    | 8 | 9 | 20 | 0 | 15 | 9 | 8 | **486** |

*Table 2 Legend: Error Categories By Frequency*

| 0 | 1-5 | 6-10 | 11-15 | 16-25 |
|---|-----|------|-------|-------|

The confusion matrix from the inner circle varieties is shown in Table 3. Because the model performs better, there are fewer errors throughout. Once again, however, the presence of more errors indicates that regions are being confused which, in turn, indicates that the regional varieties involved are more similar. Note that the number of samples for the Leipzig corpora is much higher because it contains more data. The largest source of error is confusion between Australian and New Zealand English (213). The next is between New Zealand and the UK (175) followed by New Zealand and Australia (137). This matrix is not symmetrical because, for example, samples from Australia can be predicted to be from New Zealand and vice-versa. This indicates that New Zealand English, while generally distinct, is the least distinct of all these varieties and thus the most difficult to model (at least, in terms of CxG usage): 41.5% of all errors are misclassified samples from New Zealand.

*Table 3. Confusion Matrix for Classification with Leipzig Corpora*

|    | *AU* | *CA* | *NZ* | *UK* | *ZA* |
|----|------|------|------|------|------|
| *AU* | **7,261** | 91 | 137 | 96 | 9 |
| *CA* | 44 | **7,526** | 16 | 5 | 2 |
| *NZ* | 213 | 31 | **7,080** | 175 | 73 |
| *UK* | 74 | 13 | 98 | **7,421** | 42 |
| *ZA* | 5 | 1 | 28 | 32 | **7,552** |

*Table 3 Legend: Error Categories By Frequency*

| 1-10 | 11-50 | 51-100 | 101-200 | 200-250 |
|------|-------|--------|---------|---------|

### 3.5. Measuring Regional Similarity in the Aggregate

We also want to visualize the similarity between regions as predicted by the model. One approach is to base similarity on error analysis, but this is less robust given the relatively small number of errors that are made. Instead, we turn to the feature weights produced by the model. Here we take the feature space as observations and look at the weights of each feature for each region. As noted above, these feature weights are the output of the model and represent the predictive power of each construction for a specific region. Only positive feature quantifications (frequency) are used; this means that the output feature weights, which always fall between 1 and -1, indicate attraction to the relevant class: a weight of 1 indicates that a feature is highly predictive for the sample to belong to that regional variety and a weight of -1 indicates that a feature is highly predictive for the sample to not belong to that regional variety. Because cross-validation is used to evaluate classification performance, the classifier produces potentially different feature weights across folds. The feature weights used here come from training the classifier on the same data but without using cross-validation to shuffle the data segmentation.

In a model such as this, with a large number of features, the importance of any given construction is relatively small. This is illustrated in Figure 4, which shows the weights of individual features as vertical lines, so that vertical spikes indicate that a single feature has high positive or negative weight. The scale of the graph ranges from 0.1 to -0.1. Thus, even a feature which reaches the top of the graph still has a small weight. This figure shows that most of the predictive power of the model comes from regional variations in usage rather than regional differences in the inventory of constructions itself.

*Figure 4. Feature Weights by Region*

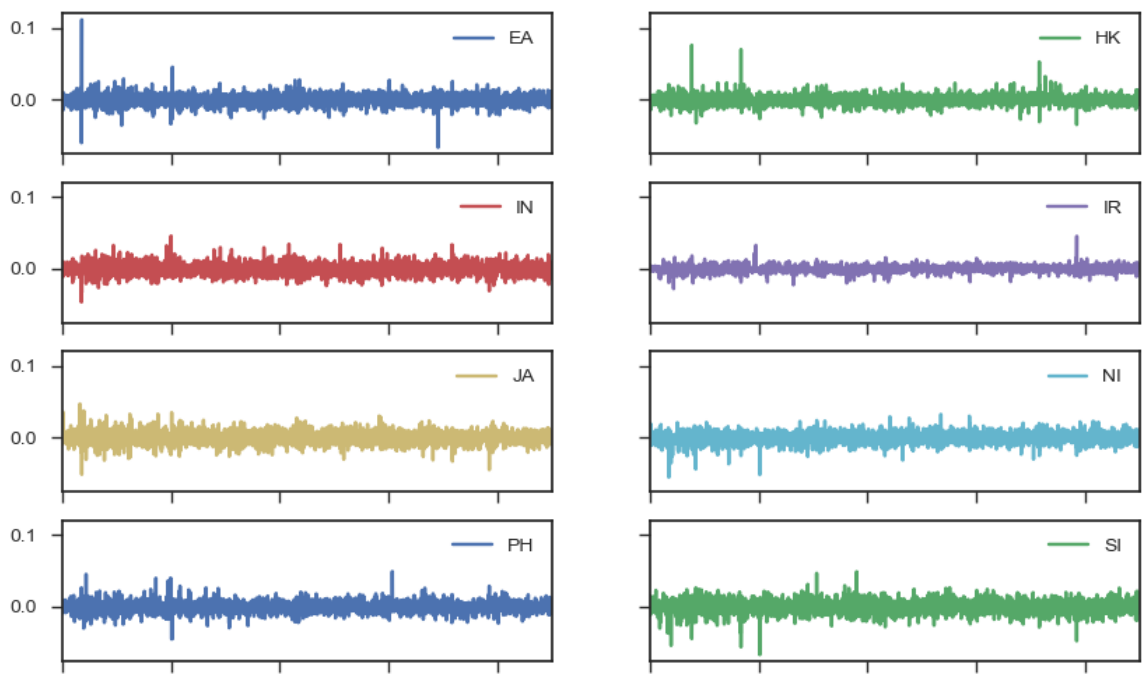

We can visualize similarity in the overall feature by applying Principal Components Analysis (PCA) to reduce the features to two dimensions and then visualizing regions as points in a two-dimensional space. To apply PCA here, we take individual features as columns and a feature's weight for a region as row (so that the ICE model has eight rows). No rotation is used. The resulting components have the explained variance shown in Table 4; in both cases additional components would increase the overall explained variance but would reduce our ability to visualize the similarity between these regional varieties.

*Table 4. Explained Variance for PCA Dimensions*

|  | Dimension 1 | Dimension 2 | Total |
|---|---|---|---|
| ICE Corpora | 20.9% | 18.9% | 39.8% |
| Leipzig Corpora | 30.9% | 29.4% | 60.3% |

*Figure 5. Similarity of Feature Weights, ICE*

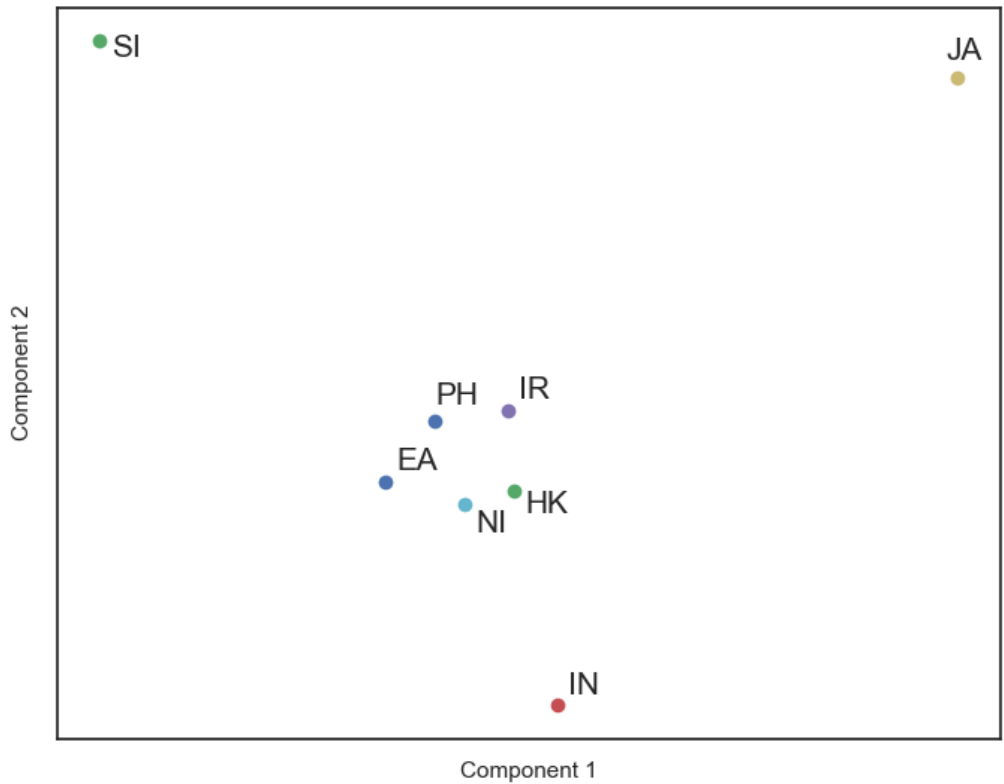

Similarities for outer circle varieties are shown in Figure 5. Jamaican English and Singapore English are the clear outliers and Irish English, interestingly, is a central point to which other varieties are equally distant in different directions. This is perhaps reflective of the fact that Irish English is an older variety representing the sort of regional variety from which all others emerged. It is useful to include it in this model as a point of comparison. We also see the importance of different components (although more components could be included). For example, while Singapore English and Jamaican English are quite distant on Component 1, they are nearly identical on Component 2.

For inner circle varieties in Figure 6, the closest varieties are Canadian English and South African English, on both dimensions. New Zealand is clearly separated from UK English on both dimensions; while New Zealand and Australia are separated on Component 2, they are the most similar to one another on Component 1. Here, Canadian English takes the more central position, being relatively equidistant from all other varieties.

*Figure 6. Similarity of Feature Weights, Leipzig*

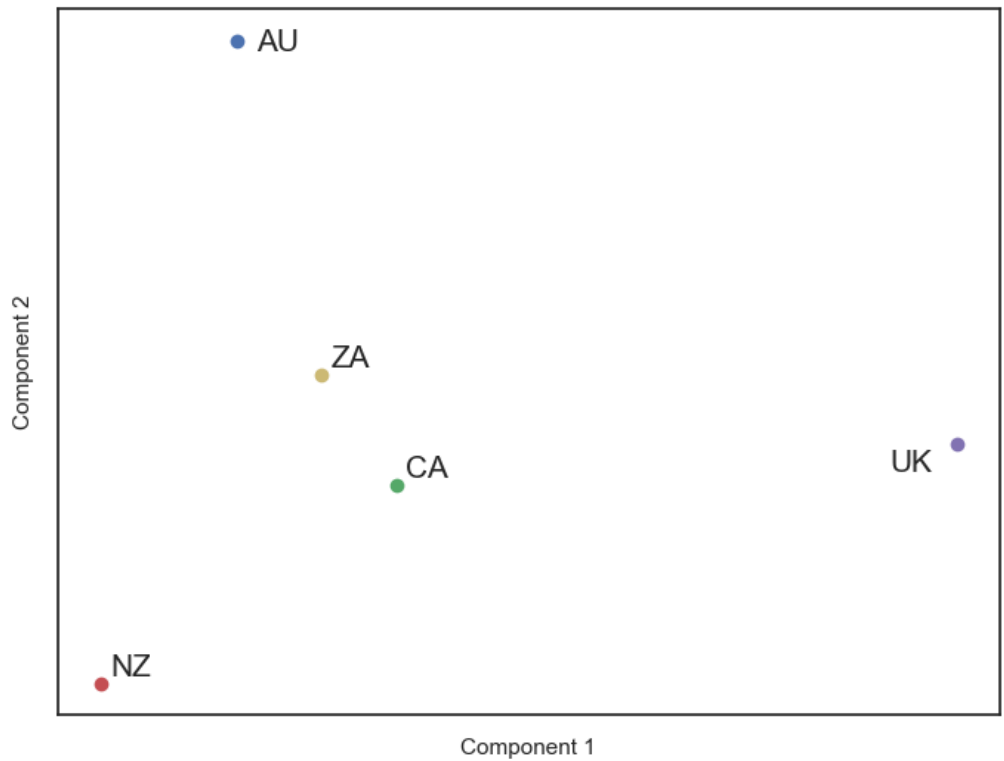

The similarity relationships in Figures 5 and 6 represent a single time period. Without diachronic data we cannot know whether this similarity is stable or whether some varieties are converging or diverging. As a result, it is difficult to evaluate alternate explanations for these similarities. For example, one line of explanation is that other major varieties like American English are influencing these varieties: perhaps Nigerian English is becoming more like American English but Philippines English is being influenced instead by Singapore English. Over time, this would be marked by increasing similarity between these pairs of regional varieties. Without diachronic corpora, it could just as easily be the case that these similar varieties are growing less similar over time: perhaps Philippines English is actually being influenced by American English and diverging from Singapore English. The point here is that this approach to dialectometry offers a way to observe such changes but these particular datasets do not.

### 3.6. Measuring the Spatial Conditioning of Individual Constructions

The next task is to examine the degree to which individual constructions are spatially conditioned by quantifying their attraction to specific regions. This is a matter of predictive power: a construction that is useful for predicting one region over another is spatially conditioned by that region to the same degree that it is predictive. More predictive features are more spatially conditioned. In the extreme, a construction used in only a single region would have perfect predictive power for that region and a construction used equally in all regions would have no predictive power at all. We can use a feature's weight from the classifier model to measure its predictive power (i.e., to make a selection signature for each region).

A full sample of constructions with regional selection signatures is given in the resources accompanying this paper and a small selection is given in Appendix 1. In general, the conditioning of any given construction is relatively small and it is the combination of constructions that provides the model its predictive power. In this section we focus on general properties of the types of constructions selected by the different regions by taking the top 250 constructions for each region and providing descriptive overviews in Figure 7 (by construction length) and Figure 8 (by slot constraints).

*Figure 7. Regional Constructions By Length*

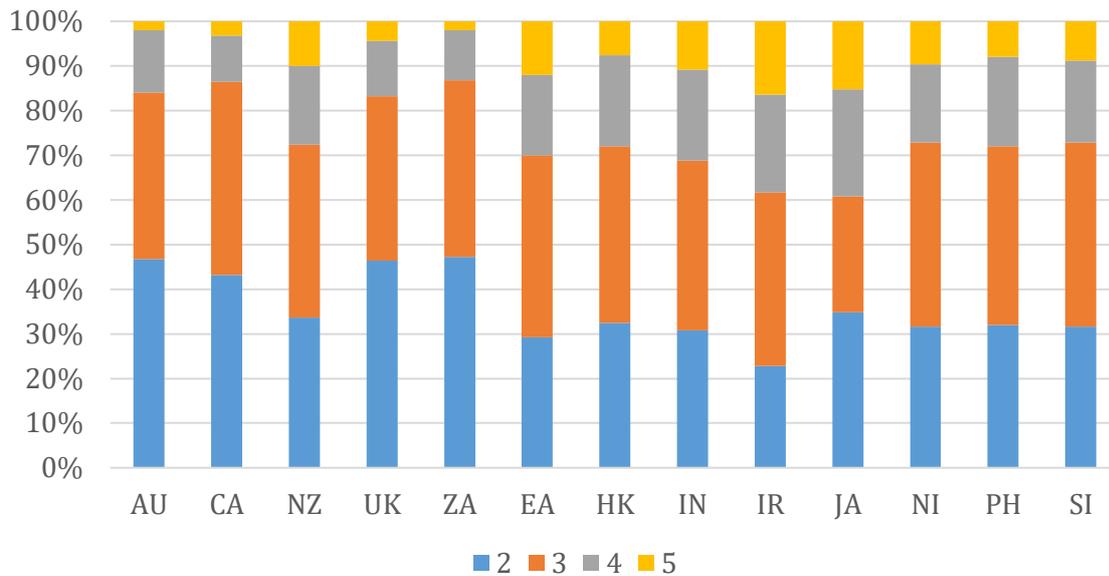

In Figure 7 we see the relative breakdown of the top constructions for each region by length, with the total bars reaching 100%. The regions from ICE favor longer constructions than the inner circle varieties from web-crawled corpora. Strictly speaking, this variation falls across registers and we are unable to say whether this is a regional or register-based variation. It is also the case, however, that New Zealand English favors longer constructions to the same degree as the outer circle varieties, indicating that this may be a regional property. The regions have more similar inventories of 3-slot and 4-slot constructions, with the real source of variation lying in the prominence of 5-slot constructions. It is also the case that longer constructions are less frequent given samples of fixed size, so that shorter constructions may be more predictive simply because they are more likely to occur.

In Figure 8 we see a similar breakdown of the type of representation used to define slot constraints. The same constructions are available to each variety, so this is an issue of which sub-set of the grammar is most predictive of each variety. Unlike construction length, there is no clear divide between registers. As before, syntactic representations dominate, but we also see significant variation across regions. For example, South African English has an equal distribution of representation types, while Irish English has more syntactic and fewer lexical representations.

*Figure 8. Regional Constructions By Slot Constraints*

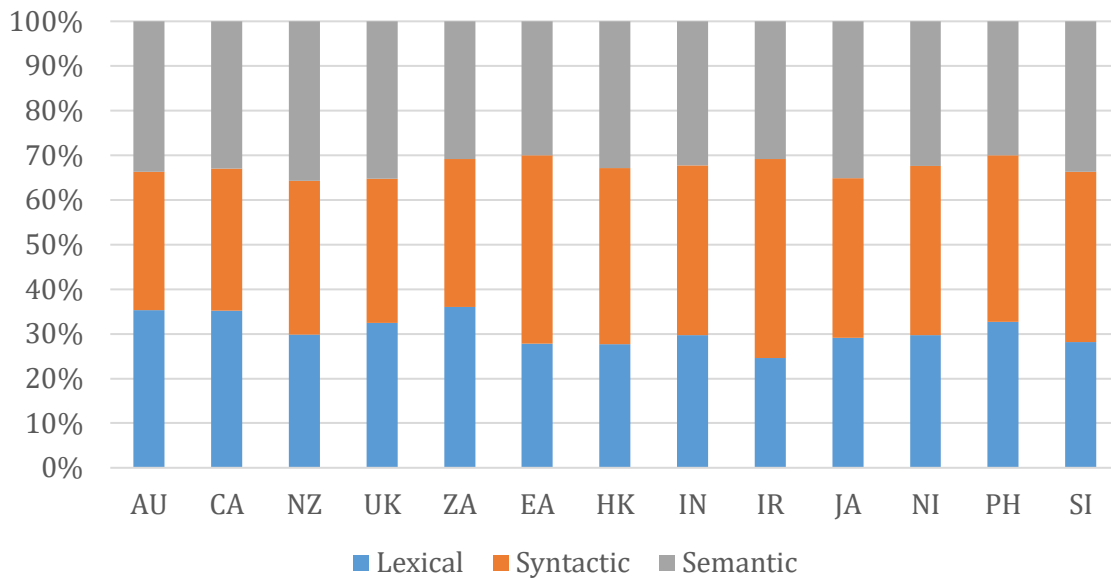

The purpose of Figures 7 and 8 is to look at variations in the types of constructions preferred by regional varieties of English. This provides a more meaningful representation than lists of predictive constructions, as in Appendix 1, because the performance of each model is based on a large number of constructions that are subject to small amounts of variation. This approach allows us to see variations in regional CxGs without over-interpreting small differences in the weights of individual constructions.

## 4. Evaluating Regional Varieties

The final task is to evaluate whether each of the regions in these two datasets forms a linguistically independent region. The classification approach requires that we build geographic assumptions into the model: how does this starting assumption influence the analysis? For example, do East Africa and Nigeria actually represent independent varieties or would the model perform better by combining these regions into a single variety? To answer this, a hierarchical ensemble classifier is used to evaluate the starting set of regions.

This classifier, based on a Linear SVM, attempts to combine similar regions into a single class by first computing a matrix of pairwise classification accuracies and then attempting to merge the pair with the lowest F-Measure (i.e., the most similar regions). This combined class is then under-sampled in order to ensure it is balanced with other regions and the performance of this new division of the corpus is evaluated on a held-out test set. The merger is accepted if it improves the overall model accuracy. The point of this algorithm is to evaluate whether the corpus as divided into regions provides the optimum set of varieties, with the caveat that this evaluation does not redraw the boundaries of each region but only alters the set of regions. The segmentation of the data into training and testing sets is again handled using cross-validation.

In this case, no mergers are possible in either dataset that improve the overall validity of the model as measured using its F-Measure on testing data. Thus, this counter-factual evaluation shows that the geographic assumptions of this study are defendable. This is expected, in this case, because these regional varieties are known and widely studied. However, it is important to have a data-driven methodology in place to test these assumptions.

## 5. Conclusions

This paper has argued that both dialectometry and cognitive sociolinguistics undertake to capture the overall variation in construction usage although they discuss this task in very different terms. For dialectometry, the problem is to model as many variants as possible. For cognitive sociolinguistics, the problem is to model as much of the functional space for expressing meaning as possible. In both cases, this paper has shown that a corpus-based approach to regional CxGs is capable of producing high-quality models of regional variation validated against ground-truth predictions.

These findings are important for cognitive linguistics because they show that the sorts of variations in usage that CxG expects to find can be modelled in a reproducible and falsifiable manner given corpus data. On the one hand, we know that there are extensive individual differences in grammar and usage (i.e., Dąbrowska, 2012, 2014). On the other hand, we now know that there are extensive collective differences in grammar and usage across groups of individuals. Both of these sources of variation are important for our understanding of usage-based grammar and the mechanisms by which constructions become entrenched both cognitively and socially.

## Acknowledgements

This research was supported in part by an appointment to the Visiting Scientist Fellowship at the National Geospatial-Intelligence Agency administered by the Oak Ridge Institute for Science and Education through an interagency agreement between the U.S. Department of Energy and NGA. The views expressed in this presentation are the author's and do not imply endorsement by the DoD or the NGA.

**Appendix 1: Spatially-Conditioned Constructions**

This appendix contains five of the top constructions for each region. The models ultimately depend on a large number of constructions, each of which has a relatively small degree of conditioning. A small number of highly predictive features for a region indicates a shallow model that is exploiting some irregularity in a small number of samples from that region (c.f., Koppel, et al., 2007). Thus, these top features only include those with a feature weight less than 0.02, a threshold that removes a very small number of unusually predictive features that occur infrequently. In order to aid interpretation of these representations, examples of the semantic domains contained here are given in Appendix 2.

*East Africa*

[<25> – ADV – 'that']
['one' – <25> – PRON]
['out' – 'of']
['one' – PRON]
[<25> – 'from' – NOUN]

*Hong Kong*

[PRON – VERB – PRON – NOUN]
['government' – noun]
[NOUN – NOUN – 'is']
[DET – 'world']
['do' – <25> – VERB]

*India*

[VERB – PRON – 'is']
[ADP – PRON – PRON – VERB]
[<25> – VERB – 'there']
[ADP – <25> – <25> – 'this']
[AUX – 'given' – <25>]

*Ireland*

[''s – VERB]
[<25> – 'and' – PRON – AUX]
[''s – <25> – ADP]
['say' – <25>]
['said' – PRON]

*Jamaica*

[<25> – SCONJ – <25> – ADV]
['end' – 'of']
[<25> – 'in' – NOUN – ADP]
['would' – VERB – <25> – <25> – <25>]
[ADP – 'a' – <25> – <25> – DET]

*Singapore*

[VERB – 'down']
['my' – ADJ]
[DET – VERB – ADV]
[DET – <25> – 'as']
['when' – 'the']

*Australia*

['people' – ADP]
[<25> – 'young' – NOUN]
[<47> – CONJ]
['use' – 'of']
[AUX – 'only']

*Canada*

['please' – VERB]
['all' – ADP]
[<49> – NOUN – <25>]
['for' – ADJ – NOUN – ADP]
['it' – VERB – DET]

*New Zealand*

['high' – <25>]
[<25> – 'required' – <25>]
[<49> – AUX]
['you' – 'to']
['or' – ADP – DET]

*United Kingdom*

['are' – VERB – <25> – <25> – <25>]
['taken' – ADP]
['down' – <25>]
[<25> – 'this' – VERB]
['range' – ADP]

*Nigeria*

[NOUN – <96>]
[SCONJ – 'are']
[NOUN – 'from' – <25>]
['of' – 'and']
[ADP – 'people']

*South Africa*

['you' – 'to']
[DET – 'world']
[<25> – <39> – <25>]
['where' – PRON – <25>]
['your' – ADJ]

*Philippines*

['and' – NOUN – CONJ]
[<25> – 'let']
[SCONJ – <25> – VERB – PRON]
['that' – <25> – <25> – ADV – <25>]
[ADP – 'other' – NOUN]

## Appendix 2: Examples of Semantic Domains

This appendix shows 10 lexical items that belong to each of a select number of semantic domains, selected to aid interpretation of the example representations in Appendix 1. A complete inventory of each semantic domain is contained in the external resources accompanying this paper.

| **<25>** | **<39>** | **<47>** |
|---|---|---|
| auditorium | wheelchairs | law |
| industry | contraband | concurrence |
| fundraisers | yard | severally |
| members | spare | exempts |
| press | depots | sentence |
| delighted | handpicked | federal |
| appeared | storage | purporting |
| wondered | assortment | administering |
| expecting | wheelie | certifying |
| discovering | torches | commissioners |

| **<49>** | **<96>** |
|---|---|
| srt | occupations |
| cetls | government-sponsored |
| aba | homebuy |
| rcr | anti-poverty |
| cmg | burglary |
| gnn | self-build |
| lcs | householder |
| gdl | landfill |
| pss | dwellers |
| ecc | municipal |